%% file: arxiv.tex
\documentclass[sigconf,nonacm]{acmart}
\makeatletter
\global\@ACM@balancefalse
\makeatother

\AtBeginDocument{%
  }

\usepackage{array}
\usepackage{colortbl}
\usepackage{enumitem}
\usepackage{multirow}
\usepackage{tcolorbox}
\usepackage{listings}
\usepackage{multicol}
\usepackage{needspace}
\input{appendix/memory_prompt_style}

\usepackage{tabularx}
\usepackage{amsmath}
\usepackage{xspace}
\usepackage{tikz}
\usepackage{supertabular}
\usepackage{booktabs}      
\usepackage{xcolor} 

\definecolor{TableAccent}{RGB}{230, 240, 255}
\usetikzlibrary{arrows.meta,positioning}

\IfFileExists{libertine.sty}{}{\microtypesetup{expansion=false}}

\newcommand{\method}{Auditable Evidence Memory\xspace}
\newcommand{\benchmark}{civic Web data benchmark\xspace}

\definecolor{TableBlue}{HTML}{145B75}
\definecolor{TableTeal}{HTML}{256D75}
\newcommand{\tablegroup}[1]{\textcolor{TableTeal}{\textit{#1}}}
\newcommand{\tableemph}[1]{\textcolor{TableBlue}{\textbf{#1}}}
\newcommand{\tablestyle}{%
  \normalsize
  \setlength{\heavyrulewidth}{0.7pt}%
  \setlength{\lightrulewidth}{0.35pt}%
  \setlength{\cmidrulewidth}{0.3pt}%
  \renewcommand{\arraystretch}{1.04}%
}

\setcopyright{none}

\begin{document}

\title[CivicBench: A Benchmark for Civic Data Analytics]{CivicBench: A Benchmark and Memory Framework for Web-Based Civic Data Analytics}

\author{Yangyu Wu}
\authornote{Yangyu Wu and Siru Zhong contributed equally.}
\email{ywu494@connect.hkust-gz.edu.cn}
\affiliation{%
  \institution{The Hong Kong University of Science and Technology (Guangzhou)}
  \city{Guangzhou}
  \country{China}
}
\author{Siru Zhong}
\authornotemark[1]
\email{szhong691@connect.hkust-az.edu.cn}
\affiliation{%
  \institution{The Hong Kong University of Science and Technology (Guangzhou)}
  \city{Guangzhou}
  \country{China}
}
\author{Haoyuan Liang}
\email{liang-hy26@mails.tsinghua.edu.cn}
\affiliation{%
  \institution{Tsinghua University}
  \city{Beijing}
  \country{China}
}
\author{Yuxuan Liang}
\authornote{Corresponding author.}
\email{yuxuanliang@hkust-gz.edu.cn}
\affiliation{%
  \institution{The Hong Kong University of Science and Technology (Guangzhou)}
  \city{Guangzhou}
  \country{China}
}

\renewcommand{\shortauthors}{Wu, Zhong, Liang, and Liang}

\begin{abstract}
\input{sections/00-abstract}
\end{abstract}

\maketitle

\input{sections/01-introduction}
\input{sections/02-related-work}

\input{sections/03-benchmark}
\input{sections/04-method}
\input{sections/05-experiments}

\input{sections/06-conclusion}

\section*{Ethics and Privacy Statement}

The benchmark supports pre-decision analysis and audit, not automated policy making, enforcement, eligibility, or risk scoring. It uses public 311 reports, one NOAA station for city-level windows, and area-weighted ACS aggregates; TSM retains bounded current-task projections. Provenance makes source, scope, and support inspectable, while retained data may increase provider exposure.

\bibliographystyle{ACM-Reference-Format}
\bibliography{references}

\appendix
\input{sections/07-supplementary}

\end{document}

%% file: appendix/memory_prompt_style.tex
\definecolor{basepurple}{RGB}{102, 51, 153}
\definecolor{memcyan}{RGB}{0, 128, 128}
\definecolor{sectionblue}{RGB}{0, 102, 153}
\definecolor{taskgreen}{RGB}{34, 139, 34}
\definecolor{outputorange}{RGB}{204, 102, 0}
\definecolor{rubricgray}{RGB}{80, 80, 80}
\definecolor{codebg}{RGB}{248, 248, 252}
\definecolor{bordergray}{RGB}{180, 180, 190}
\definecolor{sharedbg}{RGB}{250, 250, 252}
\definecolor{memhighlight}{RGB}{230, 248, 255}

\tcbuselibrary{skins,breakable}

\newtcolorbox{sharebox}{
    enhanced, breakable, colback=sharedbg, colframe=rubricgray!30,
    boxrule=0.5pt, arc=2pt,
    left=6pt,right=6pt,top=4pt,bottom=4pt,
    fontupper=\normalsize\sffamily
}

\newtcolorbox{basebox}{
    enhanced, breakable, colback=basepurple!5, colframe=basepurple!30,
    boxrule=0.5pt, arc=2pt,
    left=6pt,right=6pt,top=4pt,bottom=4pt,
    fontupper=\normalsize\sffamily\textcolor{basepurple!70!black}
}

\newtcolorbox{membox}{
    enhanced, breakable, colback=memhighlight, colframe=memcyan!40,
    boxrule=0.5pt, arc=2pt,
    left=6pt,right=6pt,top=4pt,bottom=4pt,
    fontupper=\normalsize\sffamily\textcolor{memcyan!80!black}
}

\newtcolorbox{scopebox}{
    enhanced, breakable, colback=outputorange!4, colframe=outputorange!28,
    boxrule=0.5pt, arc=2pt,
    left=6pt,right=6pt,top=4pt,bottom=4pt,
    fontupper=\normalsize\sffamily\textcolor{rubricgray}
}

\newtcolorbox{promptframe}{
    enhanced, breakable,
    colback=white, colframe=bordergray,
    boxrule=1pt, arc=3pt,
    left=12pt,right=12pt,top=10pt,bottom=10pt,
    fontupper=\normalsize\sffamily,
    toptitle=0pt,bottomtitle=0pt
}

\lstdefinestyle{promptcode}{
    backgroundcolor=\color{codebg},
    basicstyle=\ttfamily\small,
    breaklines=true,
    frame=single,
    rulecolor=\color{bordergray},
    keywordstyle=\color{sectionblue}\bfseries,
    stringstyle=\color{taskgreen},
    showstringspaces=false,
    columns=fullflexible,
    keepspaces=true,
    xleftmargin=8pt,xrightmargin=8pt,
    aboveskip=3pt,belowskip=3pt
}

\lstdefinestyle{jsoncode}{
    backgroundcolor=\color{codebg},
    basicstyle=\ttfamily\small,
    breaklines=true,
    frame=single,
    rulecolor=\color{bordergray},
    keywordstyle=\color{memcyan}\bfseries,
    stringstyle=\color{taskgreen},
    showstringspaces=false,
    columns=fullflexible,
    keepspaces=true,
    xleftmargin=10pt,xrightmargin=10pt,
    aboveskip=3pt,belowskip=3pt
}

%% file: sections/00-abstract.tex
Municipalities publish service requests, geographic boundaries, weather observations, demographic estimates, and operational guidance through independently maintained Web portals. Civic analysis must combine these sources while preserving their schemas, source links, spatial and temporal scopes, and interpretive limits. Tool-using language agents can perform this integration, but existing task benchmarks rarely test whether a computed claim remains supported by the records used. We present CivicBench, a 400-task benchmark over four cities and two years that materializes civic Web data into frozen bundles for reproducible evaluation. Executable contracts score computation, source linkage, coverage, scope, qualified answers, and refusal when the requested conclusion is not supported. We also introduce Task-State Memory (TSM), which compiles tool observations into typed assertions, reconciles records only within compatible scopes, and retrieves a bounded state organized around unresolved constraints. Across nine completed model pairs, eight yield positive mean-score differences with 95\% semantic-contract-cluster bootstrap intervals excluding zero; the ninth interval crosses zero. The benchmark and memory framework make the records behind a claim inspectable at the point of decision.

%% file: sections/01-introduction.tex
\section{Introduction}

\begin{figure}[t]
  \centering
  \includegraphics[width=\columnwidth]{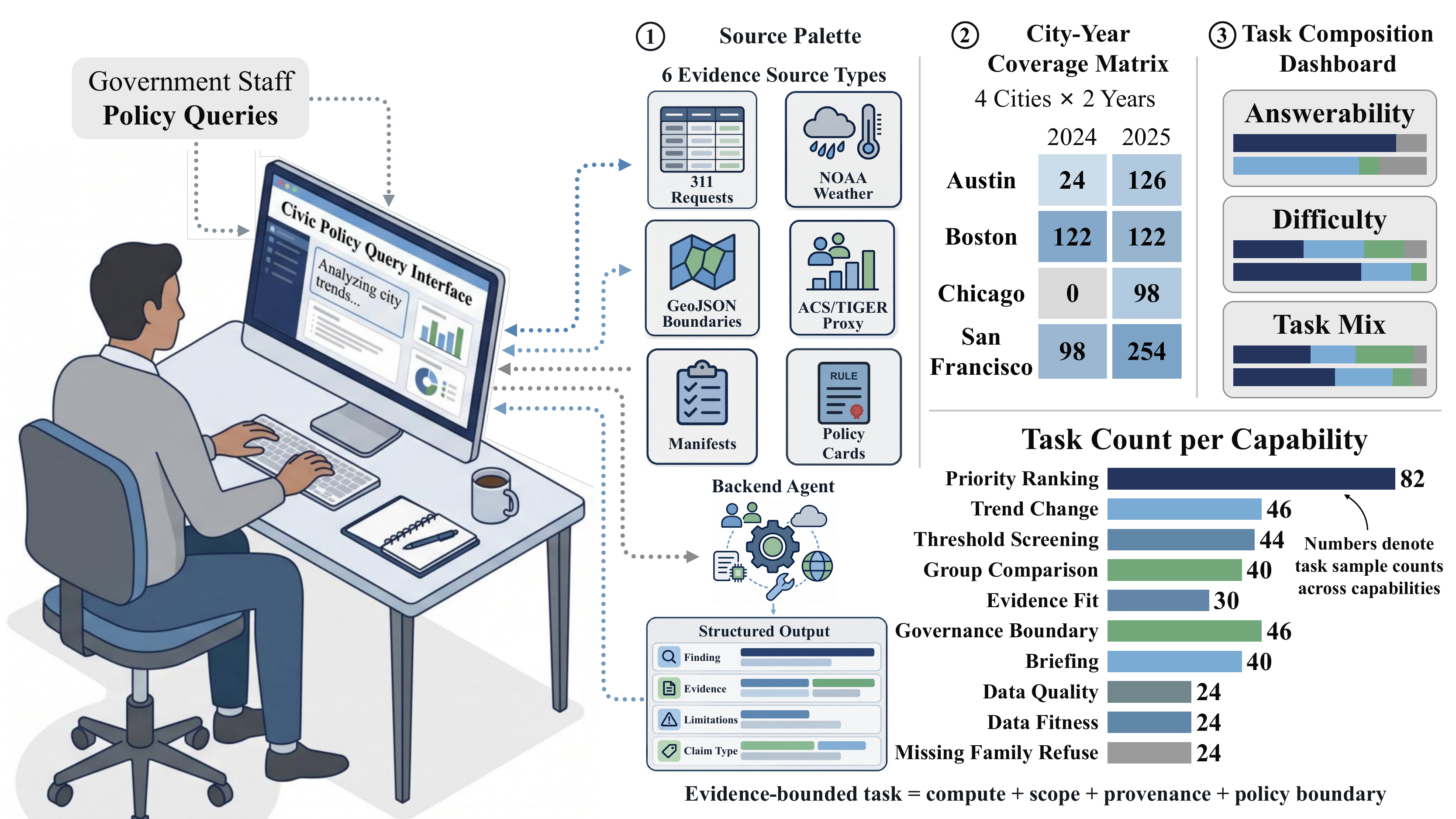}
  \caption{Benchmark scope and task composition. Six public evidence sources cover four cities and two years; each of the 400 tasks combines a computation with provenance, scope, and a claim boundary.}
  \Description{Six civic Web evidence sources feed a benchmark of 400 tasks across four cities, two years, ten capabilities, and three answer types.}
  \label{fig:overview}
\end{figure}

Civic Web portals combine municipal service records with federal weather, boundary, and demographic data for ranking, screening, and oversight analysis \cite{wang2017signature,agonafir2022streetflooding,kontokosta2021urbaninformatics}. Their public availability does not make them interchangeable: independently maintained portals differ in schema, identifiers, coverage, spatial units, update cycles, and evidentiary meaning \cite{yu2012open,white2018promises,kontokosta2021bias}. Civic analysis is therefore a Web data integration problem in which provenance and quality determine the claims that can be made. We freeze portal-derived records into identical local bundles, retaining publisher, identifier, schema, and scope metadata. This setting isolates cross-source evidence use from live-site and interface drift while preserving the boundaries that a valid claim must carry.

Tool-using language-model agents fit this setting because they decompose open-ended objectives, invoke specialized tools for external evidence and exact computation, and adapt to intermediate observations \cite{yao2023react,schick2023toolformer}. The evaluation challenge is to distinguish a numerically plausible answer from a claim whose computation, provenance, coverage, scope, and claim type are all supported. The agent must also preserve those support links across requirements, observations, failed attempts, and unresolved subgoals. Recent work describes this failure as \emph{behavioral state decay} and treats memory as an active intervention that surfaces decision-relevant state when it matters \cite{wu2026remembermatters}. Civic analysis adds a domain invariant: retained state must remain attributable and correctly scoped. Complaint counts measure reported activity; station rainfall defines a city-level observation window; and an area poverty rate is a descriptive aggregate rather than an individual attribute or causal variable \cite{white2018promises,kontokosta2021bias,liu2023underreporting,robinson1950ecological}. We call the resulting requirement \emph{scope-aware agent reasoning}: compute the strongest result supported by observed evidence and stop at its boundary.

Existing Web and tool-agent benchmarks emphasize interaction success or task completion in Web and software environments \cite{zhou2024webarena,xie2024osworld,liu2024agentbench,mialon2024gaia,yao2024taubench}. Data-analysis benchmarks make database and notebook workflows increasingly realistic \cite{li2023bird,lei2025spider2,hu2024infiagent}, with most tasks beginning from an integrated database or workspace. Agent-memory benchmarks and systems study conversational recall, long-term interaction, or reusable experience \cite{packer2024memgpt,gutierrez2024hipporag,maharana2024locomo,wu2025longmemeval,wang2025awm}. The missing evaluation target is the joint use of heterogeneous civic Web data and scope boundaries. The missing memory property is reconstructability: retained evidence should expose its provenance, scope, and conflicts when an agent forms a claim.

We address these gaps with a benchmark and a memory mechanism. The \benchmark contains 400 tasks over four cities and two years, spanning ten capabilities in ranking, comparison, screening, data-fitness assessment, briefing, and scope-aware refusal. Each task pairs a structured computation with an analysis brief and an executable contract that checks the result, provenance, coverage, scope, and claim type. \method (TSM) maintains typed evidence instead of a free-form transcript: deterministic \textsc{Write} compiles observations with provenance and scope, conservative \textsc{Resolve} reconciles compatible assertions, and bounded \textsc{Retrieve} selects evidence for unresolved constraints. The benchmark tests adherence to Web scope boundaries, while TSM keeps those boundaries reconstructable across tool use (\autoref{fig:overview}). Our study makes three contributions:

\begin{itemize}[leftmargin=*]
  \item \textbf{An scope-aware civic Web data benchmark.} We construct and review 400 reproducible tasks over four cities and two years from frozen portal evidence, with executable measurements, explicit spatial and temporal scopes, answer contracts, contract-preserving frames, and evidence-gated scoring.
  \item \textbf{Task-State Memory.} Deterministic typed \textsc{Write}, same-key and same-scope \textsc{Resolve}, and bounded constraint-centered \textsc{Retrieve} preserve provenance, conflicts, and claim boundaries through a compact agent/tool interface.
  \item \textbf{Cross-model evaluation and diagnostics.} Across nine completed model pairs, eight have positive score differences with 95\% cluster-bootstrap intervals excluding zero. Frozen reference suites support module ablations plus traceability, failure, and resource analyses that expose provenance, scope, reliability, and efficiency trade-offs.
\end{itemize}

Together, the benchmark and TSM shift evaluation from Web task completion or memory recall to scope-aware record use: recovering support for a computed claim while preserving its provenance, scope, and limitations. The resulting task design, failure taxonomy, and controlled memory analysis connect Web data integration with record-aware agent planning in public-sector workflows.

%% file: sections/02-related-work.tex
\section{Related Work}

\subsection{Web and Data-Agent Benchmarks}

Web and general agent benchmarks test browser interaction, computer use, tool calling, and long-horizon completion \cite{zhou2024webarena,xie2024osworld,liu2024agentbench,mialon2024gaia,patil2025bfcl,yao2024taubench}. Their primary unit is interaction or task completion in an interface; our unit is a computed claim whose support must remain traceable to correctly scoped evidence from multiple Web publishers. BIRD, Spider~2.0, and InfiAgent-DABench increase schema, computation, and workflow realism \cite{li2023bird,lei2025spider2,hu2024infiagent}, while most tasks begin from an integrated database or workspace and score the answer or execution. Attribution benchmarks test support from cited passages, and unanswerability benchmarks test whether models refrain when evidence is insufficient \cite{gao2023alce,kirichenko2025abstention}. The civic setting requires both properties for claims computed across heterogeneous structured Web sources. Our benchmark freezes those sources for reproducibility and jointly evaluates tool execution, computation, provenance, spatial and temporal scope, qualified answers, and evidence-grounded refusal.

\subsection{Memory in Language-Model Agents}

Agent memory has been framed as context paging, associative retrieval, temporal knowledge graphs, learned note organization, experience retention, or reusable workflows \cite{packer2024memgpt,kang2025memoryos,gutierrez2024hipporag,xu2025amem,rasmussen2025zep,zhong2024memorybank,wang2025awm}. Proactive-memory work treats memory as an intervention for recovering decision-relevant state from long trajectories \cite{wu2026remembermatters}, while LoCoMo, LongMemEval, and MemoryAgentBench measure recovery from interaction histories \cite{maharana2024locomo,wu2025longmemeval,hu2025memoryagentbench}. An evidence ledger adds three invariants: each assertion remains traceable to a tool observation, differently scoped statements coexist, and unresolved conflicts remain visible to downstream reasoning. Our method is closest in spirit to typed and provenance-aware memory systems \cite{zhang2025gmemory,sumers2024coala} and deterministic truth-maintenance principles \cite{dekleer1986atms}. Its operational focus is public evidence roles and unresolved task constraints: tool outputs become typed assertions with file, field, spatial, temporal, and step provenance, and retrieval projects that state into a bounded context whose claims remain reconstructable.

\subsection{Civic Web Data and Urban Agents}

Open-government publication expands access to civic data but does not establish common semantics, analytical fitness, or policy transparency \cite{yu2012open}. UrbanGPT studies spatiotemporal forecasting \cite{li2024urbangpt}; CityBench and USTBench evaluate broad urban capabilities and spatiotemporal reasoning by urban agents \cite{feng2025citybench,lai2025ustbench}; GeoLLM and LLM-Find study geospatial knowledge and GIS data retrieval \cite{manvi2024geollm,ning2025llmfind}; and surveys organize the emerging urban-agent literature \cite{han2025urbanagent}. These lines emphasize prediction, spatial knowledge, reasoning, or discovery. Our benchmark evaluates analysis across independently published civic Web sources, where source meaning and scope determine valid conclusions. Municipal requests measure reported activity and administrative processes \cite{white2018promises,kontokosta2021bias,liu2023underreporting}, while spatial aggregation introduces ecological and modifiable-areal-unit risks \cite{robinson1950ecological,openshaw1984maup}. We encode these constraints as executable evidence contracts, explicit scopes, and evidence-gated answer types.

%% file: sections/03-benchmark.tex
\section{Civic Web Data Benchmark}
\label{sec:benchmark}

Each task combines a governance-facing question $q$, a city's two-year evidence bundle $E$, public scope $C^{\mathrm{pub}}$, and response schema $A$. All tasks for a city receive the same evidence bundle. Hidden executable specifications, evidence paths, and reference-answer provenance support deterministic scoring but are never exposed to the agent.

\begin{figure*}[t]
  \centering
  \includegraphics[width=\textwidth]{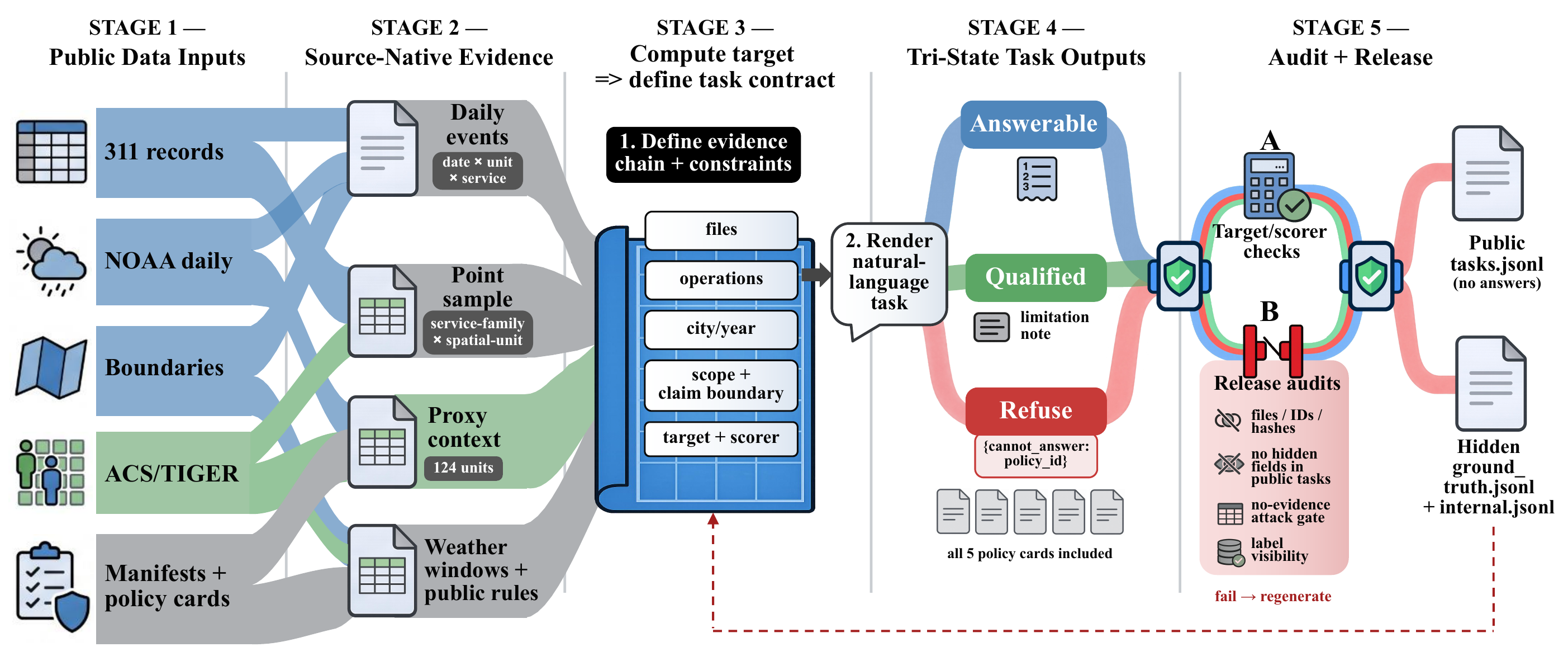}
  \caption{Benchmark construction pipeline. Evidence-derived targets and contracts are fixed before governance-facing questions are framed. Public tasks, evaluator ground truth, and internal generation state remain separate, while target/scorer, operation, and no-record shortcut checks gate the frozen artifact.}
  \Description{A five-stage pipeline acquires public civic data, constructs source-native evidence, derives targets and contracts, frames tasks, and checks the frozen benchmark.}
  \label{fig:benchmark-construction}
\end{figure*}

Figure~\ref{fig:benchmark-construction} summarizes the pipeline from public inputs to source-native evidence, contract-first target construction, governance-facing task framing, review, and artifact freezing.

\subsection{Data and Evidence Assets}

\paragraph{Snapshot registration.}
We build eight full-calendar-year snapshots from public Web portals for Austin, Boston, Chicago, and San Francisco over 2024--2025. A source configuration fixes the city and date range, upstream portal dataset and station identifiers, source taxonomy, and weather thresholds. After acquisition, a city-year manifest records retrieval time, observed coverage, row counts, materialized paths, and source limitations. Table~\ref{tab:data-sources} identifies the upstream municipal, boundary, and weather records. ACS 2024 five-year estimates and TIGER/Line tract geometry are shared across cities to construct descriptive population and socioeconomic context. Freezing these sources removes live-site and update drift while retaining the publisher, identifier, schema, and scope metadata required for cross-source reasoning.

\input{tables/benchmark_data_sources}

\paragraph{Traceable transformation.}
Municipal Web records are mapped to four common service families while preserving each city's declared administrative or operational units. For every city-year, we then materialize two answer-free tables: (i) daily events indexed by date, unit, and service family, with request counts, closure aggregates, and full-record coordinate coverage; and (ii) a deterministic sample of at most 20 valid-coordinate records per service-family--unit cell. Across the eight snapshots, these assets contain 195,769 daily rows and 15,550 sampled point rows. The daily table supports exact temporal and coverage computations, while the point sample supports reproducible boundary diagnostics without distributing every coordinate.

\paragraph{Two-year evidence packaging.}
Each city has a fixed 16-file bundle containing daily tables, point samples, NOAA files, a boundary GeoJSON, manifests, shared proxy metadata, and policy cards; every city task receives the same bundle. Public tasks provide scope, response schema, and measurement definitions, while source files contain no targets. Definitions require observed weather triggers and valid baselines, exclude invalid spatial labels from rankings while retaining them for citywide accounting, treat absent valid samples as unknown, and fix eligible sets and grouping rules for population-normalized comparisons.

\subsection{Contract-First Task Construction}

Task construction begins after the source assets are fixed:
\begin{enumerate}[leftmargin=*,nosep,label=\arabic*.]
  \item The generator derives an eligible city, period, and service family from the source metadata and pairs that scope with a capability and requested-claim specification.
  \item It executes the capability-specific operation over the source assets to obtain the structured reference result and its coverage.
  \item Before the question is written, it freezes an executable contract containing the operation, complete constraints, answerability, output fields, scorer, and target. An source chain records the required source roles and operations, while oracle provenance records how the target was derived.
  \item It renders the fixed contract through one of four public-sector frames (operational handoff, planning decision, oversight review, or decision deliverable) without changing the underlying computation or claim boundary.
  \item It emits three ID-aligned records: a public task, hidden evaluator ground truth, and an internal generation record. Release checks then verify their separation and freeze the tasks, evidence, policies, and manifests by hash.
\end{enumerate}
The public record contains the task, scope, declared files, response schema, measurement definitions, and policy constraints. Hidden ground truth retains the reference answer, scoring contract, source chain, and oracle provenance; the internal record reconstructs generation.

\subsection{Benchmark Composition and Answerability}

Table~\ref{tab:tasks} summarizes the 400-task evaluation set; no training or development split is released. Austin, Boston, Chicago, and San Francisco contribute 98, 116, 98, and 88 tasks, respectively. The set contains 400 unique questions, 316 semantic specifications, and 305 distinct structured answers. Seventy-two specifications have two or three scenario views, covering 156 tasks. These repeated views support framing-consistency analysis, and our uncertainty estimation clusters them by their shared semantic specification instead of treating each view as independent. The four difficulty bands denote required reasoning behavior and are not human-difficulty ratings.

\input{tables/benchmark_task_composition}

Each contract has one of three outcomes. A \emph{direct answer} is supported at the requested scope. A \emph{qualified answer} returns the supported computation while explicitly limiting a broader interpretation. An \emph{evidence-scope-aware refusal} identifies a missing category or temporal grain and returns the finest alternative supported by the available data. The benchmark tests whether the agent can recover the values, coverage, scope, and supported alternative from evidence.

Programmatic benchmark generation can mistake surface variation for substantive construct coverage \cite{bean2025construct}. Construction checks verify that every question preserves its city, period, service, and operation anchors and that repeated scenario views share an identical contract. Question-only answerability prediction is reported as a visibility diagnostic; answer recovery still requires evidence and computation.

\subsection{Evaluation Protocol}

\begin{figure}[ht]
  \centering
  \includegraphics[width=\columnwidth]{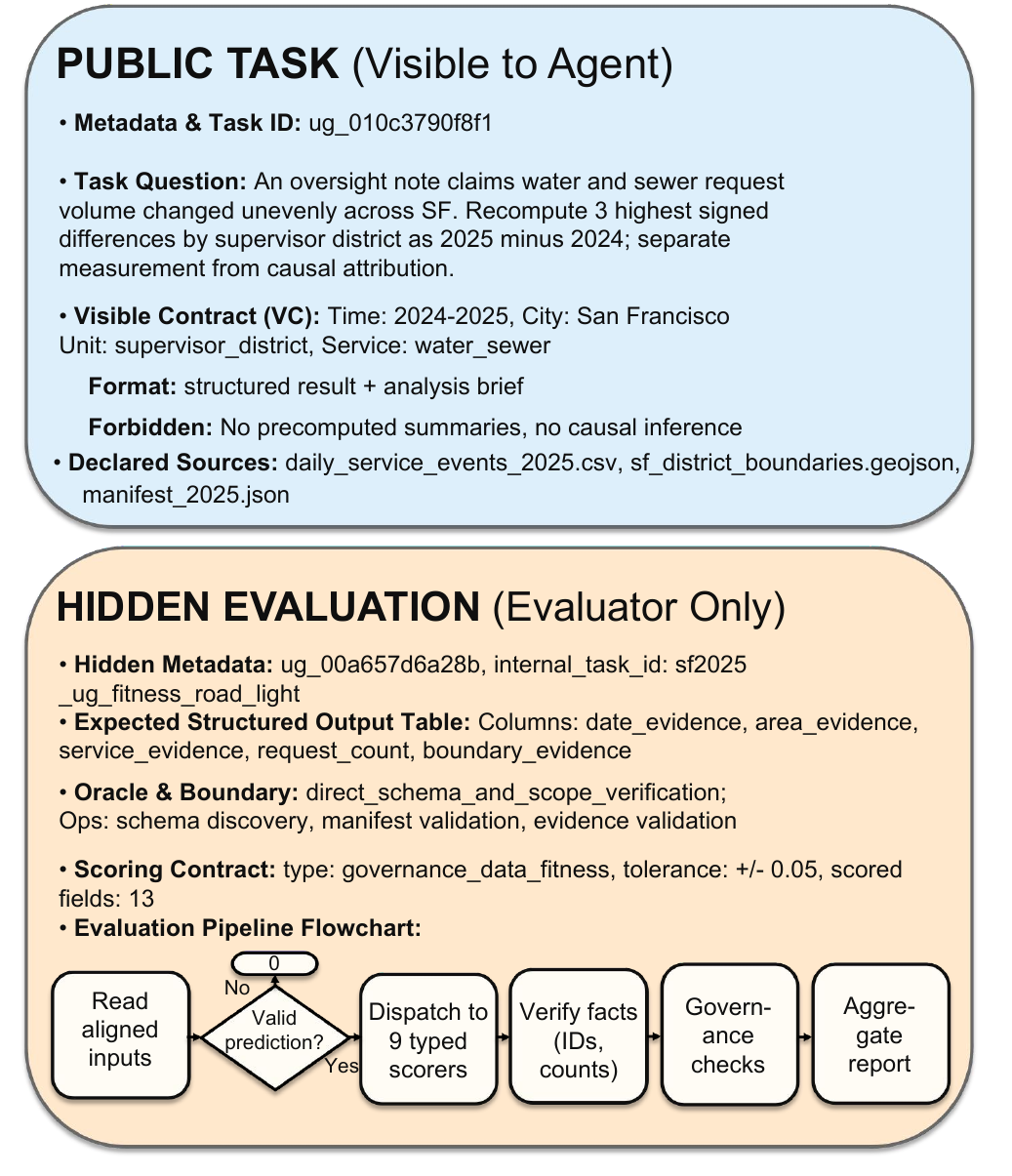}
  \caption{Public task and hidden evaluation. The agent receives the question, visible contract, and declared source files. Scoring type, oracle provenance, source traces, and evaluator state remain hidden until prediction is complete.}
  \Description{The public task exposes a question, contract, and source files to the agent, while the scoring contract, oracle provenance, and evaluator remain hidden.}
  \label{fig:data-case}
\end{figure}

At evaluation time, an agent receives one public task record and tool access restricted to its 16 declared source files. It returns a task-specific \texttt{structured\_result} and an \texttt{analysis\_brief} with finding, evidence, limitations, and claim type. Figure~\ref{fig:data-case} shows this split: the public record exposes the question, visible contract, and allowed files, while the hidden scoring contract, oracle, and evaluation pipeline remain outside the model context. The hidden contract dispatches an operation-specific scorer that checks required sets, orderings, numeric values, provenance and coverage fields, claim boundary, or supported alternative. Numeric checks use fixed tolerances. The brief contributes deterministic completeness and exact claim-type checks; open-ended writing quality is outside the score.

Each scorer returns a task score $s_i\in[0,1]$ and a strict exact-match indicator. The primary score is the unweighted mean across 400 tasks, reported with exact match and breakdowns by capability, answerability, difficulty, and city. Missing predictions, malformed outputs, and failed runs score zero; tool use, latency, and tokens are logged separately.

Boundary credit follows the underlying evidence. A qualified response earns it after recovering the supporting computation; a refusal must identify a verifiable evidence gap and return the finest supported alternative. Valid empty results report the evaluated units and coverage. This design separates evidence-based qualification from generic caution, keyword stuffing, and always-refuse behavior.

Before release, blocking checks verify artifact separation, source-record integrity, target/scorer closure, contract-preserving framing, and key operation semantics recomputed from source evidence. Five no-record strategies are evaluated overall and by capability and operation. These checks establish internal consistency and resistance to the tested shortcuts. Broader validity studies and legal review remain necessary for public redistribution.

%% file: tables/benchmark_data_sources.tex
\begin{table}[t]
  \caption{Civic Web sources and two-year service-record snapshots.}
  \label{tab:data-sources}
  \centering
  \tablestyle
  \setlength{\tabcolsep}{3pt}
  \begin{tabularx}{\columnwidth}{@{}>{\raggedright\arraybackslash}p{0.15\columnwidth}>{\raggedright\arraybackslash}Xr@{}}
    \toprule
    City & Municipal, boundary, and NOAA sources & 311 rows \\
    \midrule
    Austin & Austin Open Data 311 (\texttt{xwdj-i9he}); council districts (\texttt{w3v2-cj58}); Camp Mabry (\texttt{USW00013958}) & 280,395 \\
    \addlinespace[1.5pt]
    Boston & Analyze Boston 311; BPDA neighborhood boundaries; Logan (\texttt{USW00014739}) & 324,124 \\
    \addlinespace[1.5pt]
    Chicago & Chicago Data Portal 311 (\texttt{v6vf-nfxy}); community areas (\texttt{igwz-8jzy}); O'Hare (\texttt{USW00094846}) & 445,933 \\
    \addlinespace[1.5pt]
    San Francisco & SF Open Data 311 (\texttt{vw6y-z8j6}); supervisor districts (\texttt{hcgx-vtsb}); Downtown (\texttt{USW00023272}) & 1,454,388 \\
    \midrule
    \textbf{Total} & & \textbf{2,504,840} \\
    \bottomrule
  \end{tabularx}
  \smallskip
  \begin{minipage}{\columnwidth}
    \footnotesize\raggedright
    \textit{Note.} Identifiers follow the release metadata. Counts are retained service-request rows; NOAA contributes daily station observations, and ACS/TIGER contributes a shared 124-unit proxy table.
  \end{minipage}
\end{table}

%% file: tables/benchmark_task_composition.tex
\begin{table*}[t]
  \caption{Composition of the 400-task benchmark by evidence operation and answer type. D3 contributes 254 tasks that require explicit qualification.}
  \label{tab:tasks}
  \centering
  \tablestyle\small
  \setlength{\tabcolsep}{2.5pt}
  \renewcommand{\arraystretch}{1.02}
  \begin{tabularx}{\textwidth}{@{}c>{\raggedright\arraybackslash}p{0.145\textwidth}>{\raggedright\arraybackslash}X>{\raggedright\arraybackslash}p{0.295\textwidth}lr@{}}
    \toprule
    Band & Capability & Required evidence operation & Expected output & Answer & $n$ \\
    \midrule
    \textbf{D1} & Data fitness & Check schema, manifest scope, and category presence & Field- and file-level evidence & Direct & 24 \\
    \midrule
    \multirow{3}{*}{\textbf{D2}}
      & Priority ranking & Aggregate a scoped metric; rank eligible units & Ranked units with metric and coverage & Direct & 30 \\
      & Threshold screening & Align dates; build event/baseline windows; threshold units & Selected units with window statistics & Direct & 44 \\
      & Data quality & Audit coordinate coverage and sampled boundary fit & Spatial-risk ranking and citywide accounting & Direct & 24 \\
    \midrule
    \multirow{6}{*}{\textbf{D3}}
      & Priority ranking & Aggregate a scoped metric; rank eligible units & Ranked units with metric and coverage & Qualified & 52 \\
      & Trend change & Align 2024--2025 records; compute signed changes & Ranked changes with interpretation boundary & Qualified & 46 \\
      & Group comparison & Join proxies; normalize metrics; compare fixed groups & Group statistics and descriptive contrast & Qualified & 40 \\
      & Evidence fit & Test whether spatial evidence supports the target product & Fitness verdict and supported fallback & Qualified & 30 \\
      & Governance boundary & Compute descriptive change; test the governance claim & Supported result plus rejected attribution & Qualified & 46 \\
      & Briefing & Synthesize evidence with reporting/weather limitations & Finding, evidence, and explicit limitations & Qualified & 40 \\
    \midrule
    \textbf{D4} & Missing family/grain & Check category and temporal-grain availability & Refusal with finest supported alternative & Refusal & 24 \\
    \midrule
    \multicolumn{5}{@{}l}{\textbf{Total}\quad 122 direct; 254 qualified; 24 refusal} & \textbf{400} \\
    \bottomrule
  \end{tabularx}
  \smallskip
  \begin{minipage}{\textwidth}
    \footnotesize\raggedright
    \textit{Bands.} D1 validates evidence; D2 derives deterministic results; D3 adds explicit interpretation boundaries; D4 verifies evidence gaps and returns the finest supported alternative.
  \end{minipage}
\end{table*}

%% file: sections/04-method.tex
\section{Task-State Memory}
\label{sec:method}

\begin{figure*}[t]
  \centering
  \includegraphics[width=\textwidth]{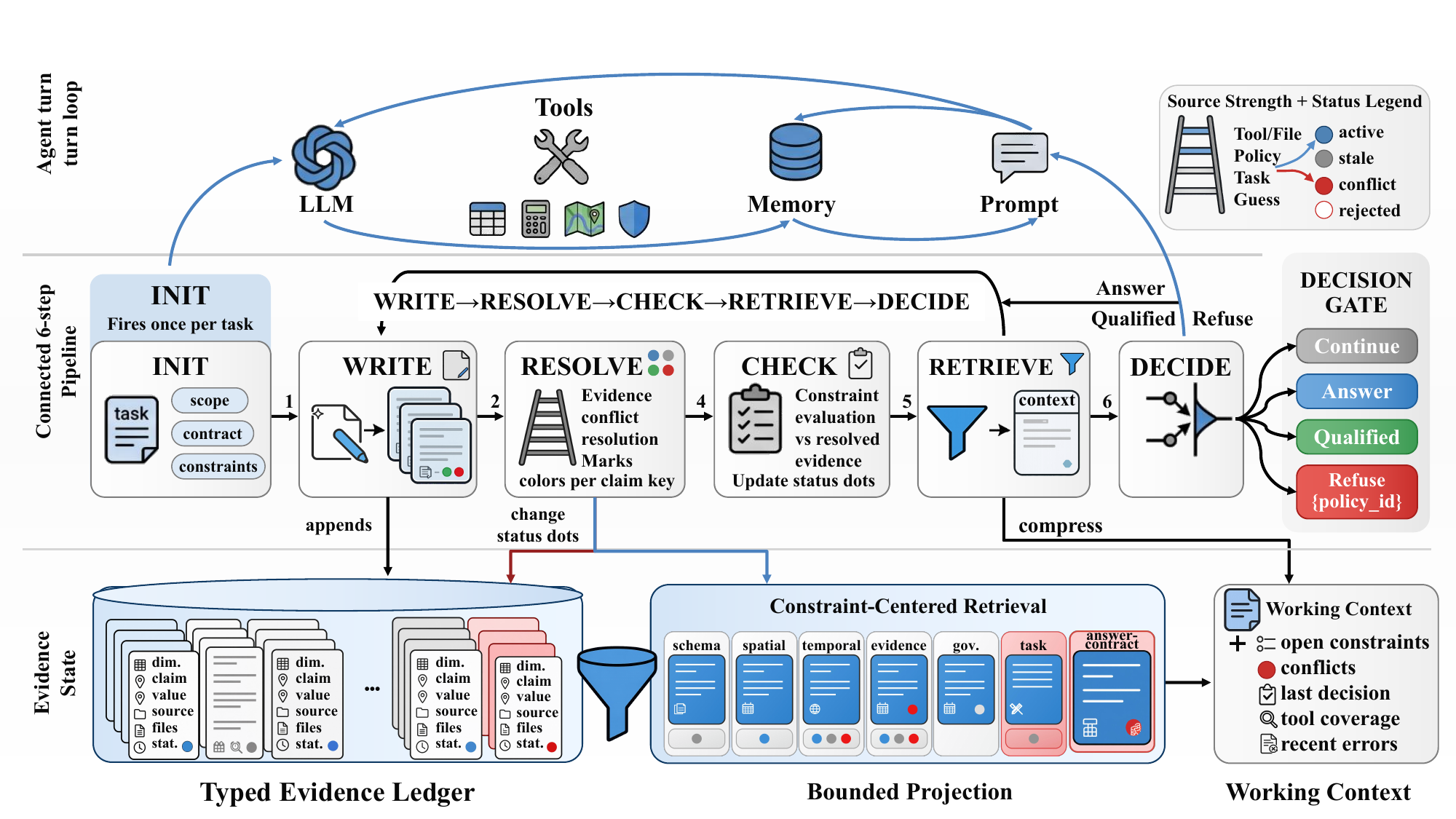}
  \caption{Task-State Memory. INIT compiles the public task; WRITE, RESOLVE, and CHECK maintain typed records and constraint state; RETRIEVE forms a bounded projection; and DECIDE uses that context to continue, answer, qualify, or refuse.}
  \Description{A six-stage memory loop compiles a task, writes and resolves typed observations, checks constraints, retrieves a bounded projection, and supports the model's next decision.}
  \label{fig:method}
\end{figure*}

Task-State Memory (TSM) is an external, typed state for agents integrating heterogeneous civic Web records. As shown in Figure~\ref{fig:method}, \textsc{Init} compiles the task once. Each tool observation is written, resolved, checked, and retrieved before the next \textsc{Decide}; the decision continues the loop or emits a terminal response. The first five stages are deterministic, while \textsc{Decide} remains the model's choice. The separation keeps the task state reconstructable and keeps free-form transcript text outside the memory state.

\subsection{Six-Stage Memory Loop}

\paragraph{INIT: Compile the public contract.}
Let $P(\mathcal{T})$ denote the canonical public task projection. Baseline and memory conditions receive the same exogenous input,
\begin{equation}
  X_B^0=X_M^0=P(\mathcal{T}).
\end{equation}
TSM compiles the task intent, answer contract, declared file roles, and spatial and temporal requirements into typed task assertions and verifiable constraints. Initialization performs no file I/O. A declared path records an allowed input, whereas file contents enter the task state after a successful tool call. The distinction keeps task metadata separate from data-supported support.

\paragraph{WRITE: Compile observations into typed evidence.}
After action $a_t$ returns serialized observation $o_t$, a deterministic compiler updates the ledger,
\begin{equation}
  L_t'=f_{\mathrm{write}}(L_{t-1},P(\mathcal{T}),a_t,o_t).
\end{equation}
Each assertion has the form
\begin{equation}
e_i=(d_i,c_i,v_i,s_i,F_i,K_i,\sigma_i^S,\sigma_i^T,p_i,z_i,t_i),
\end{equation}
where $d_i$ is its dimension, $c_i$ and $v_i$ its claim and value, $s_i$ its source, $F_i$ and $K_i$ its files and fields, $\sigma_i^S$ and $\sigma_i^T$ its spatial and temporal scope, $p_i$ its provenance parents, $z_i$ its status, and $t_i$ its creation step. Dimensions cover schema, evidence, spatial, temporal, governance, task, and answer-contract roles. The compiler derives schema observations, computation results, file and field provenance, observed scopes, policy evidence, evidence gaps, and tool errors. Computations retain a fingerprint of the complete result when the prompt shows a preview. Failed tool executions receive the rejected status and cannot support a claim; a successful but irrelevant call does not satisfy provenance or computation constraints. Equal public inputs and serialized observations therefore produce identical ledger updates.

\paragraph{RESOLVE: Reconcile only compatible assertions.}
TSM compares a new assertion with prior assertions that share its normalized claim key and identical spatial and temporal scope. Identical observations reaffirm the stronger or newer assertion; compatible schema and scope fragments merge; and a stronger observed source supersedes a weaker duplicate. Equal-strength contradictions remain conflicts, while differently scoped facts coexist. The ledger is append-only: superseded assertions become stale, and failed observations remain rejected. Its prompt-facing status classes are active, stale, conflict, and rejected. This policy keeps a city-, year-, or neighborhood-specific fact separate from evidence at another scope.

\paragraph{CHECK: Evaluate evidence against task constraints.}
TSM recomputes checks over the typed ledger: required schemas and manifests must be observed, cited evidence must come from declared files, computations must use required source roles, and scopes must match the task. Coupled analyses require joint computation rather than separate source inspections. Each check stores supporting and blocking assertions with a satisfied, unsatisfied, conflicted, or unverified status; a missing computation cannot justify refusal.

\paragraph{RETRIEVE: Form a bounded projection.}
Once the fixed retrieval trigger is reached, TSM converts the full task state into a constraint-centered bounded projection. Eligible evidence is tool-supported and active or conflict-bearing. Selection preserves conflicts and observed boundary evidence, covers missing source roles, retains nonredundant computations, and ranks remaining assertions by open-constraint relevance and recency. The projection contains selected assertions with provenance and scope, evidence-role coverage, open or conflicted checks, recent tool errors, and a compact answer-contract reminder. It is deterministically fitted to a fixed character budget and omits evidence already present in retained tool pairs. The typed links let the model reconstruct why a claim is available, limited, or blocked.

\paragraph{DECIDE: Gate the next agent action.}
The model receives the public task, common tool interface, retained recent interaction pairs, and the projected working context. It emits one of four explicit gates: \textsc{Continue} requests another tool call; \textsc{Answer} returns a directly supported result; \textsc{Qualified} returns a supported descriptive or coarser result with its scope; and \textsc{Refuse} marks an unsupported claim when no adequate alternative can be computed. A \textsc{Continue} action produces the next observation and restarts the update loop. A terminal response is checked against the public response contract before evaluation. The gate is a model decision informed by TSM; the memory state does not supply an answer.

\subsection{Agent Interface and Comparison Conditions}

ReAct and TSM share the model, base system prompt, public task, tool schema, evidence permissions, maximum actions, response format, and evaluator. They choose actions $a_t^R$ and $a_t^A$ independently, so their trajectories can diverge. Equal actions return equal serialized results:
\begin{equation}
  a_t^R=a_t^A \Longrightarrow O(a_t^R)=O(a_t^A),
\end{equation}
where $O$ is the common tool environment and equality is over the serialized raw result. Full-trace ReAct retains every request/result pair. TSM retains the two latest raw pairs and receives the bounded typed projection described above; its ablations remove one workflow stage while preserving the remaining interface and execution budget. For RQ3, Raw Retrieval is an independently executed memory condition with the same model, tasks, two-pair history, injection trigger, and character budget. It retains raw observations instead of typed writes and projections, supporting evidence-role coverage comparison without assertion-level provenance checks.
\input{tables/results_main_comparison}

%% file: tables/results_main_comparison.tex
\begin{table*}[ht]
  \caption{RQ1 effectiveness across nine model pairs; eight TSM intervals exclude zero.}
  \label{tab:main}
  \centering
  \small
  \setlength{\tabcolsep}{4.5pt}
  \renewcommand{\arraystretch}{1.14}
  \begin{tabular*}{\textwidth}{@{\extracolsep{\fill}}l r >{\columncolor{TableAccent}}r r r >{\columncolor{TableAccent}}r r >{\columncolor{TableAccent}}r@{}}
    \toprule
    & \multicolumn{3}{c}{\textbf{Mean task score} $\uparrow$} & \multicolumn{2}{c}{\textbf{Exact match} $\uparrow$} & \multicolumn{2}{c}{\textbf{Completion} $\uparrow$} \\
    \cmidrule(lr){2-4}\cmidrule(lr){5-6}\cmidrule(lr){7-8}
    \textbf{Model} & \textbf{ReAct} & \cellcolor{TableAccent}\textbf{TSM} & \textbf{$\Delta$ [95\% CI]} & \textbf{ReAct} & \cellcolor{TableAccent}\textbf{TSM} & \textbf{ReAct} & \cellcolor{TableAccent}\textbf{TSM} \\
    \midrule
    \multicolumn{8}{l}{\textit{Frozen reference runs}} \\
    DeepSeek V4 Flash & 0.2782 & 0.3451 & \textbf{+0.0669 [0.0336, 0.1008]} & 0.0000 & 0.0050 & 0.9425 & 0.9500 \\
    Grok 4 Fast Reasoning & 0.2502 & 0.3102 & \textbf{+0.0600 [0.0263, 0.0936]} & 0.0100 & 0.0150 & 0.9950 & 0.9700 \\
    \addlinespace[2pt]
    \multicolumn{8}{l}{\textit{Additional endpoints}} \\
    GPT-5.4-nano & 0.1373 & 0.2089 & \textbf{+0.0717 [0.0473, 0.0962]} & 0.0025 & 0.0000 & 0.9950 & 0.9550 \\
    Gemini-3.5-flash-lite & 0.0974 & 0.1506 & \textbf{+0.0532 [0.0241, 0.0831]} & 0.0075 & 0.0075 & 0.3775 & 0.4550 \\
    Qwen3.6-35B-A3B & 0.1090 & 0.1598 & \textbf{+0.0508 [0.0303, 0.0723]} & 0.0000 & 0.0000 & 0.4150 & 0.3800 \\
    GLM-5.2 & 0.1644 & 0.2010 & \textbf{+0.0366 [0.0097, 0.0638]} & 0.0000 & 0.0000 & 0.5225 & 0.3975 \\
    Kimi-K2.7-Code & 0.3825 & 0.4590 & \textbf{+0.0766 [0.0426, 0.1097]} & 0.0000 & 0.0050 & 0.9950 & 0.8925 \\
    GPT-5.6-terra & 0.5414 & 0.5577 & +0.0163 [-0.0050, 0.0383] & 0.0000 & 0.0000 & 1.0000 & 1.0000 \\
    Claude Sonnet 5 & 0.3395 & 0.4334 & \textbf{+0.0938 [0.0606, 0.1273]} & 0.0075 & 0.0000 & 0.6950 & 0.7875 \\
    \bottomrule
  \end{tabular*}
  \smallskip
  \begin{minipage}{\textwidth}
    \scriptsize\raggedright
    \textit{Note.} Intervals use 20,000 semantic-contract-cluster bootstrap resamples; bold intervals exclude zero. Blue cells identify TSM values and do not encode outcome quality.
  \end{minipage}
\end{table*}

%% file: sections/05-experiments.tex
\section{Experiments}
\label{sec:experiments}

The benchmark measures task-level analytic performance, while the memory study measures how retained task state affects tool use, intermediate records, and final responses.

The evaluation addresses four research questions:
\begin{enumerate}[label=\textbf{RQ\arabic*:}, leftmargin=*]
  \item How does TSM affect scope-aware task performance relative to full-trace ReAct? (Sec.~\ref{sec:rq1})
  \item How does each TSM stage contribute to end-to-end performance? (Sec.~\ref{sec:rq2})
  \item Does TSM preserve correct source traces while covering the source roles and unresolved constraints required by each task? (Sec.~\ref{sec:rq3})
  \item What failures does TSM correct or introduce, and what reliability and resource trade-offs does it incur? (Sec.~\ref{sec:rq4})
\end{enumerate}

\subsection{Experimental Setup}

\paragraph{Benchmark.}
We evaluate on the fixed 400-task civic Web data test set, which spans four cities, the 2024--2025 periods, and ten capability categories. Each task exposes its declared bundle of 16 frozen files and is scored by a deterministic task-specific evaluator. All conditions receive the same task text, file bundle, tools, and response schema.

\paragraph{Models.}
The RQ1 matrix contains nine completed model pairs. Frozen runs of DeepSeek V4 Flash and Grok 4 Fast Reasoning provide the primary paired comparisons. The broader matrix includes GPT-5.4-nano, Gemini-3.5-flash-lite, Qwen3.6-35B-A3B, GLM-5.2, Kimi-K2.7-Code, GPT-5.6-terra, and Claude Sonnet 5. Every model passed a JSON-action preflight, and the artifacts record registry identifiers, access dates, runtime hashes, task selections, and run signatures.

\paragraph{Execution protocol.}
All conditions use temperature 0 and a 120-second request timeout. They allow at most eight agent steps and four API retries, with identical tool serialization and the same terminal response contract. Missing predictions, malformed finals, exhausted retries, and step exhaustion score zero. Frozen manifests record task and runtime hashes, git revision, profile, task IDs, shards, and run signatures. Merged rows must pass exact 400-task coverage and identity checks.

\paragraph{Evaluation and statistics.}
The primary metric is mean task score over 400 tasks, accompanied by exact match, completion, component failures, and resource use. We draw 20,000 paired bootstrap samples and cluster by the 316 semantic contracts, grouping repeated scenario views in each resample. The supplementary material reports pre-execution slices and benchmark-integrity diagnostics.

\subsection{RQ1: Task Performance}
\label{sec:rq1}

\paragraph{Setting.}
We compare full-trace ReAct with TSM on the same 400 tasks for each model. ReAct retains every request/result pair without an external task state. TSM retains the two most recent pairs and, after the second tool call, injects a typed task-state projection capped at 6,000 characters. The comparison measures an unbounded raw trace against bounded structured state; it is not context-length matched. We report mean task score, exact match, completion, and paired score differences with 95\% semantic-contract-cluster bootstrap intervals.

\paragraph{Results and analysis.}
Table~\ref{tab:main} shows positive TSM score differences for eight of nine model pairs, with 95\% cluster-bootstrap intervals excluding zero. In the reference runs, TSM improves mean score by 0.0669 for DeepSeek V4 Flash (95\% CI [0.0336, 0.1008]) and 0.0600 for Grok 4 Fast Reasoning ([0.0263, 0.0936]). GPT-5.6-terra is the exception, with an interval crossing zero. Completion rises from 0.9425 to 0.9500 for DeepSeek but falls from 0.9950 to 0.9700 for Grok, whose TSM run contains ten retry-exhausted provider failures. The score gains therefore do not imply a uniform completion gain.

\subsection{RQ2: Component Contribution}
\label{sec:rq2}

\paragraph{Setting.}
We use a separately frozen DeepSeek V4 Flash suite and rerun all 400 tasks for each natural-trajectory ablation, with two retained history pairs and the same eight-step budget. The four variants remove reconciliation, constraint states and check-driven ranking, task-state projection, or early terminal decisions, respectively. Each condition independently chooses its tool trajectory. The full-TSM row is the reference for this suite and is reported separately from the frozen RQ1 run.

\paragraph{Results and analysis.}
Table~\ref{tab:ablation} shows a $-0.0016$ change without \textsc{Resolve}, compared with declines of 0.0122 and 0.0223 without \textsc{Check} and \textsc{Decide}. Removing \textsc{Retrieve} causes the largest decline ($-0.1278$): a maintained ledger provides little benefit when its bounded projection is hidden from the model. Retrieval is the dominant measured contributor in this run, with smaller complementary effects from checking and decision control.

\input{tables/results_mechanism_ablation}

\subsection{RQ3: Trace Quality}
\label{sec:rq3}

\paragraph{Setting.}
We review the final TSM projection from the two 400-task RQ1 runs. Each eligible assertion is checked against a successful tool result, a read file, a field in that result, and the source's spatial and temporal scope. We also measure required evidence-role coverage over the 380 tasks that define roles and count whether constraints open in the full ledger remain visible in the projection. Raw Retrieval is an independently executed comparison with the same model, tasks, two-pair history, injection timing, and 6,000-character budget. Its raw observations support role-coverage comparison but contain no assertion-level trace links.

\paragraph{Results and analysis.}
Table~\ref{tab:tracequality} reports valid counts over eligible assertions. Every eligible assertion passes the tool-result, field, spatial, and temporal checks. File-read tracing passes for all Grok assertions and 1426/1429 DeepSeek assertions; the three failures are catalog listings mislabeled as direct reads. TSM raises required-role coverage over Raw Retrieval from 0.8395 to 0.9044 for DeepSeek and from 0.6982 to 0.7373 for Grok, while exposing 150/158 and 249/249 constraints open in the full ledger. In the 40-task provenance stress test, TSM scores 0.4074 versus 0.3782 for ReAct, and 38/40 runs record a stale or rejected assertion. These results establish high trace correctness and improved task-evidence coverage while locating a specific provenance-labeling defect.

\input{tables/results_auditability}

\subsection{RQ4: Failure Patterns}
\label{sec:rq4}

\paragraph{Setting.}
The aggregate analysis uses the frozen DeepSeek and Grok RQ1 pairs, which have complete reliability and resource artifacts. We separate exhausted API retries, terminal-tool errors, step limits, and malformed finals from non-exclusive completed-answer failures in computation, provenance and scope, and boundary or alternative handling. Resource values are unconditional means over all 400 tasks.

\paragraph{Results and analysis.}
Table~\ref{tab:reliability-cost} summarizes the paired operational changes; Supplementary Table~\ref{tab:supp-reliability-cost} gives endpoint values. TSM reduces computation failures from 277 to 262 for DeepSeek and from 337 to 296 for Grok. DeepSeek uses 6,555 fewer total tokens per task ($-11.4\%$), whereas Grok uses 515 more ($+1.6\%$). Both TSM runs invoke more tools and add about five seconds of latency. Grok also incurs ten retry-exhausted provider failures. The supplementary paired cases show corrections and regressions together with their observed resource changes.

\input{tables/results_reliability_cost}

%% file: tables/results_mechanism_ablation.tex
\begin{table}[t]
  \caption{RQ2 TSM stage ablations on the frozen DeepSeek suite. Removing Retrieve causes the largest score loss.}
  \label{tab:ablation}
  \centering
  \small
  \setlength{\tabcolsep}{4pt}
  \renewcommand{\arraystretch}{1.15}
  \begin{tabular*}{\columnwidth}{@{\extracolsep{\fill}}lrrr@{}}
    \toprule
    \textbf{Variant} & \textbf{Score} $\uparrow$ & \textbf{$\Delta$} & \textbf{Exact} $\uparrow$ \\
    \midrule
    \textbf{Full TSM} & \textbf{0.3621} & -- & \textbf{0.005} \\
    \addlinespace[2pt]
    w/o Resolve & 0.3605 & -0.0016 & 0.000 \\
    w/o Check & 0.3499 & -0.0122 & 0.000 \\
    w/o Retrieve & 0.2343 & \textbf{-0.1278} & 0.000 \\
    w/o Decide & 0.3398 & -0.0223 & 0.005 \\
    \bottomrule
  \end{tabular*}
  \smallskip
  \begin{minipage}{\columnwidth}
    \scriptsize\raggedright
    \textit{Note.} Each variant removes one stage; $\Delta$ is relative to Full TSM within this separately frozen suite.
  \end{minipage}
\end{table}

%% file: tables/results_auditability.tex
\begin{table}[t]
  \caption{RQ3 trace validity and task-evidence coverage.}
  \label{tab:auditability}
  \small
  \centering
  \tablestyle
  \setlength{\tabcolsep}{3pt}

  \begin{tabular*}{\linewidth}{@{\extracolsep{\fill}}lrr@{}}
    \toprule
    \multicolumn{3}{@{}l}{\tablegroup{(a) Assertion-level trace checks}} \\
    Trace check & DeepSeek & Grok \\
    \midrule
    Tool-result trace & 1432/1432 & 1303/1303 \\
    File-read trace & 1426/1429 & 1302/1302 \\
    Field trace & 1333/1333 & 1255/1255 \\
    Spatial-scope match & 555/555 & 404/404 \\
    Temporal-scope match & 802/802 & 631/631 \\
    \bottomrule
  \end{tabular*}

  \vspace{0.65em}
  \begin{tabular*}{\linewidth}{@{\extracolsep{\fill}}lrrrr@{}}
    \toprule
    \multicolumn{5}{@{}l}{\tablegroup{(b) Task-level evidence coverage}} \\
    Model & Raw & TSM & $\Delta$ & Open \\
    \midrule
    DeepSeek V4 & 0.8395 & \tableemph{0.9044} & \tableemph{+0.0649} & 150/158 \\
    Grok 4 Fast & 0.6982 & \tableemph{0.7373} & \tableemph{+0.0391} & 249/249 \\
    \bottomrule
  \end{tabular*}
  \smallskip
  \begin{minipage}{\columnwidth}
    \footnotesize\raggedright
    \textit{Note.} Fractions in (a) are valid/eligible assertions. Role coverage is a macro-average over 380 applicable tasks; Open is projected/full-ledger unresolved constraints. Raw Retrieval has no typed assertion links.
  \end{minipage}
\end{table}

%% file: tables/results_reliability_cost.tex
\begin{table}[t]
  \caption{RQ4 operational changes from ReAct to TSM.}
  \label{tab:reliability-cost}
  \centering
  \small
  \setlength{\tabcolsep}{3pt}
  \renewcommand{\arraystretch}{1.15}
  \begin{tabular*}{\columnwidth}{@{\extracolsep{\fill}}lrrrr@{}}
    \toprule
    \tablehead
    \textbf{Model} & \shortstack{\textbf{Complete} $\uparrow$\\\textbf{$\Delta$ (tasks)}} & \shortstack{\textbf{Failures} $\downarrow$\\\textbf{$\Delta$}} & \shortstack{\textbf{Tokens} $\downarrow$\\\textbf{$\Delta$/task}} & \shortstack{\textbf{Latency} $\downarrow$\\\textbf{$\Delta$ (s/task)}} \\
    \midrule
    DeepSeek V4 & \textbf{+3} & \textbf{-15} & \textbf{-6,555} & +4.94 \\
    Grok 4 Fast & -10 & \textbf{-41} & +515 & +5.09 \\
    \bottomrule
  \end{tabular*}
  \smallskip
  \begin{minipage}{\columnwidth}
    \scriptsize\raggedright
    \textit{Note.} Values are TSM minus ReAct; bold marks favorable changes. Endpoint counts and token components appear in Supplementary Table~\ref{tab:supp-reliability-cost}.
  \end{minipage}
\end{table}

%% file: sections/06-conclusion.tex
\section{Conclusion}

We formulate civic Web analysis as scope-aware reasoning and introduce TSM to preserve source, field, spatial, temporal, and claim boundaries during tool use. TSM improves mean task score by 0.0669 and 0.0600 in two frozen reference runs; across nine model pairs, eight 95\% semantic-contract-cluster bootstrap intervals exclude zero. Ablation identifies bounded retrieval as the largest measured contributor, while trace checks expose three catalog-linking errors. Resource effects remain model-dependent, with lower token use for DeepSeek, higher token use for Grok, and higher latency for both. Typed, constraint-centered memory improves agent performance while keeping the records behind civic Web claims inspectable.

\paragraph{Limitations.}
Stable generated wording makes answerability visible: question-only prediction is perfect, and the five no-record attacks cover specific shortcuts. The scorer evaluates structured values, evidence, and deterministic completeness, excluding human-perceived prose and policy usefulness. Frozen files isolate Web data integration, provenance, and scope from browser navigation and source updates. Because the benchmark informed architecture selection, generalization claims remain confined to this four-city artifact. One-model, independent-trajectory ablations support association, not causal decomposition. The post-run catalog-provenance patch has an unmeasured task effect, and hosted aliases add run variance. Chicago and Boston redistribution rights remain unresolved; release must separate metadata, code, hashes, and legal aggregates from restricted source-derived artifacts.

%% file: sections/07-supplementary.tex
\section{Supplementary Materials}

These tables report benchmark audits, endpoint slices, detailed reliability and resource measurements, evidence-state checks, run stability, and paired cases supporting the main analysis. Appendix~\ref{app:memory-prompt} reproduces the baseline and Memory-Core prompt specification.

\input{tables/results_benchmark_audits}
\input{tables/supplementary_rq1_capabilities}
\input{tables/supplementary_rq1_slices}
\input{tables/supplementary_rq3_evidence_audit}
\input{tables/supplementary_paired_cases}
\input{tables/supplementary_rq4_failures_resources}
\input{tables/supplementary_run_stability}
\input{tables/supplementary_provenance_stress}

%% file: tables/results_benchmark_audits.tex
\begin{table}[ht]
  \caption{Benchmark integrity and no-evidence shortcut audits. Evidence-free strategies reach at most 0.1715 overall task score.}
  \label{tab:audit}
  \centering
  \small
  \setlength{\tabcolsep}{4pt}
  \renewcommand{\arraystretch}{1.14}
  \begin{tabular*}{\columnwidth}{@{\extracolsep{\fill}}lrr@{}}
    \toprule
    \textbf{Audit} & \textbf{Overall} & \textbf{Worst subgroup} \\
    \midrule
    \multicolumn{3}{l}{\textbf{Scorer and shortcut controls}} \\
    Oracle/scorer regression & 1.0000 & 1.0000 \\
    Always refuse & 0.0983 & 0.2000 \\
    Always empty & 0.1177 & 0.2000 \\
    Keyword stuffing & 0.0891 & 0.2000 \\
    Majority answer shape & 0.0986 & 0.2038 \\
    Public-contract canned & 0.1715 & 0.3000 \\
    \multicolumn{3}{l}{\textbf{Answerability diagnostic}} \\
    Question-only label accuracy & 1.0000 & -- \\
    \bottomrule
  \end{tabular*}
  \smallskip
  \begin{minipage}{\columnwidth}
    \scriptsize\raggedright
    \textit{Note.} Shortcut rows report task score without source evidence; the label diagnostic measures answerability visibility only.
  \end{minipage}
\end{table}

%% file: tables/supplementary_rq1_capabilities.tex
\begin{table}[ht]
  \caption{RQ1 endpoint scores by capability.}
  \label{tab:supp-capabilities}
  \centering
  \tablestyle
  \setlength{\tabcolsep}{3pt}

  \begin{tabular*}{\linewidth}{@{\extracolsep{\fill}}lrrrr@{}}
    \toprule
    & \multicolumn{2}{c}{DeepSeek} & \multicolumn{2}{c}{Grok} \\
    \cmidrule(lr){2-3}\cmidrule(lr){4-5}
    Metric & ReAct & TSM & ReAct & TSM \\
    \midrule
    \multicolumn{5}{@{}l}{\tablegroup{Computation, synthesis, and quality}} \\
    Rank & 0.4390 & \tableemph{0.5198} & 0.3818 & \tableemph{0.4535} \\
    Trend & 0.3799 & \tableemph{0.7937} & 0.2799 & \tableemph{0.5855} \\
    Threshold & 0.1403 & \tableemph{0.1687} & \tableemph{0.1472} & 0.1264 \\
    Group & 0.1314 & \tableemph{0.1437} & 0.1167 & \tableemph{0.1489} \\
    Brief & 0.3277 & \tableemph{0.3689} & 0.3011 & \tableemph{0.3484} \\
    Quality & \tableemph{0.4528} & 0.1490 & \tableemph{0.5333} & 0.3917 \\
    \addlinespace[2pt]
    \multicolumn{5}{@{}l}{\tablegroup{Evidence fitness and claim boundaries}} \\
    Evidence fit & \tableemph{0.2642} & 0.2522 & 0.2580 & \tableemph{0.2644} \\
    Boundary & 0.1873 & \tableemph{0.2458} & 0.1174 & \tableemph{0.2108} \\
    Fitness & 0.1719 & \tableemph{0.2712} & 0.2297 & \tableemph{0.2932} \\
    Refusal & 0.0717 & \tableemph{0.0838} & \tableemph{0.0526} & 0.0177 \\
    \bottomrule
    \end{tabular*}
  \smallskip
  \begin{minipage}{\columnwidth}
    \footnotesize\raggedright
    \textit{Note.} Higher scores are better; bold blue marks the stronger method within each model pair and capability.
  \end{minipage}
\end{table}

%% file: tables/supplementary_rq1_slices.tex
\begin{table}[t]
  \caption{RQ1 endpoint scores by city, answer type, and evidence-role complexity.}
  \label{tab:supp-slices}
  \centering
  \tablestyle
  \setlength{\tabcolsep}{3pt}

  \begin{tabular*}{\linewidth}{@{\extracolsep{\fill}}lrrrr@{}}
    \toprule
    & \multicolumn{2}{c}{DeepSeek} & \multicolumn{2}{c}{Grok} \\
    \cmidrule(lr){2-3}\cmidrule(lr){4-5}
    Metric & ReAct & TSM & ReAct & TSM \\
    \midrule
    \multicolumn{5}{@{}l}{\tablegroup{City}} \\
    Austin & 0.3171 & \tableemph{0.3589} & 0.3207 & \tableemph{0.4010} \\
    Boston & 0.2188 & \tableemph{0.2745} & 0.1516 & \tableemph{0.2057} \\
    Chicago & 0.2794 & \tableemph{0.3624} & 0.2873 & \tableemph{0.3074} \\
    San Francisco & 0.3117 & \tableemph{0.4034} & 0.2603 & \tableemph{0.3498} \\
    \addlinespace[2pt]
    \multicolumn{5}{@{}l}{\tablegroup{Answer type and evidence roles}} \\
    Direct & \tableemph{0.3130} & 0.2976 & 0.3051 & \tableemph{0.3499} \\
    Qualified & 0.2810 & \tableemph{0.3926} & 0.2425 & \tableemph{0.3187} \\
    Refuse & 0.0717 & \tableemph{0.0838} & \tableemph{0.0526} & 0.0177 \\
    0--1 roles & 0.2862 & \tableemph{0.3926} & 0.2473 & \tableemph{0.3508} \\
    2 roles & 0.2314 & \tableemph{0.2824} & 0.2227 & \tableemph{0.2264} \\
    3 roles & \tableemph{0.4335} & 0.2406 & \tableemph{0.4062} & 0.3652 \\
    \bottomrule
    \end{tabular*}
  \smallskip
  \begin{minipage}{\columnwidth}
    \footnotesize\raggedright
    \textit{Note.} Complexity is computed from public required evidence roles only. Bold blue marks the stronger method within each model pair and slice.
  \end{minipage}
\end{table}

%% file: tables/supplementary_rq3_evidence_audit.tex
\begin{table*}[ht]
  \caption{RQ3 evidence-state trace validity and coverage.}
  \label{tab:supp-evidence-audit}
  \centering
  \tablestyle\small
  \setlength{\tabcolsep}{4pt}
  \begin{tabular*}{\linewidth}{@{\extracolsep{\fill}}llrrrrrrr@{}}
    \toprule
    & & \multicolumn{5}{c}{Valid/eligible assertion traces} & \multicolumn{2}{c}{Task coverage} \\
    \cmidrule(lr){3-7}\cmidrule(lr){8-9}
    Model & Method & Parent & File & Field & Spatial & Temporal & Role cov. & Open constraint \\
    \midrule
    DeepSeek & TSM & 1432/1432 & 1426/1429 & 1333/1333 & 555/555 & 802/802 & 0.9044 & 150/158 \\
      & Raw & untyped & untyped & untyped & untyped & untyped & 0.8395 & untyped \\
    \addlinespace[2pt]
    Grok & TSM & 1303/1303 & 1302/1302 & 1255/1255 & 404/404 & 631/631 & 0.7373 & 249/249 \\
      & Raw & untyped & untyped & untyped & untyped & untyped & 0.6982 & untyped \\
    \bottomrule
  \end{tabular*}
  \smallskip
  {\scriptsize\raggedright\textit{Note.} Raw Retrieval has no native claim-level record; its required-role coverage is reconstructed from retained observations.\par}
\end{table*}

%% file: tables/supplementary_paired_cases.tex
\begin{table*}[ht]
  \caption{RQ4 paired corrections and regressions with observable cost changes.}
  \label{tab:supp-paired-cases}
  \centering
  \tablestyle\small
  \setlength{\tabcolsep}{4pt}
  \begin{tabular*}{\linewidth}{@{\extracolsep{\fill}}lllrrrrr@{}}
    \toprule
    & & & \multicolumn{3}{c}{Outcome change} & \multicolumn{2}{c}{Resource change} \\
    \cmidrule(lr){4-6}\cmidrule(lr){7-8}
    Model & Case & Task & Score $\Delta$ & Structured $\Delta$ & Boundary $\Delta$ & Tokens $\Delta$ & Latency $\Delta$ \\
    \midrule
    DeepSeek & Structured fix & {\texttt{ug\_23eba3fdf29f}} & \tableemph{+0.8500} & +1.0000 & +0.3333 & -10,953 & -4.31 \\
      & Regression & {\texttt{ug\_72c901dcb068}} & -0.8500 & -1.0000 & -0.3333 & +14,602 & +21.65 \\
    \addlinespace[2pt]
    Grok & Scope fix & {\texttt{ug\_42fdd9bc9fd7}} & \tableemph{+0.9075} & +1.0000 & +0.4167 & +4,284 & +2.33 \\
      & Regression & {\texttt{ug\_5b5a2f57011f}} & -0.9125 & -1.0000 & -0.4167 & +22,325 & +15.65 \\
    \bottomrule
  \end{tabular*}
  \smallskip
  {\scriptsize\raggedright\textit{Note.} The artifact contains all eight selected categories and their observable traces; the table shows four representative pairs.\par}
\end{table*}

%% file: tables/supplementary_rq4_failures_resources.tex
\begin{table}[!t]
  \caption{RQ4 endpoint reliability diagnostics and resource use.}
  \label{tab:supp-reliability-cost}
  \centering
  \tablestyle\small
  \setlength{\tabcolsep}{3pt}

  \begin{tabular*}{\linewidth}{@{\extracolsep{\fill}}lrrrr@{}}
    \toprule
    & \multicolumn{2}{c}{DeepSeek} & \multicolumn{2}{c}{Grok} \\
    \cmidrule(lr){2-3}\cmidrule(lr){4-5}
    Metric & ReAct & TSM & ReAct & TSM \\
    \midrule
    \multicolumn{5}{@{}l}{\tablegroup{Failures and answer diagnostics}} \\
    Complete & 377 & 380 & 398 & 388 \\
    API & 1 & 0 & 2 & 10 \\
    Max step & 9 & 10 & 0 & 0 \\
    Malformed & 13 & 10 & 0 & 2 \\
    Computation & 277 & 262 & 337 & 296 \\
    Prov./scope & 89 & 83 & 81 & 84 \\
    Boundary/alt. & 195 & 197 & 165 & 159 \\
    \addlinespace[2pt]
    \multicolumn{5}{@{}l}{\tablegroup{Mean resources per task}} \\
    Tools & 5.64 & 5.91 & 4.70 & 5.18 \\
    Retries & 0.005 & 0.005 & 0.250 & 0.292 \\
    Input tokens & 51,828 & 44,906 & 31,938 & 32,271 \\
    Output tokens & 5,701 & 6,068 & 1,091 & 1,273 \\
    Total tokens & 57,529 & 50,974 & 33,028 & 33,543 \\
    Latency (s) & 63.60 & 68.54 & 41.36 & 46.45 \\
    \bottomrule
    \end{tabular*}
  \smallskip
  {\scriptsize\raggedright\textit{Note.} Completed-answer error categories are non-exclusive. Resource values include all 400 tasks in each condition.\par}
\end{table}

%% file: tables/supplementary_run_stability.tex
\begin{table}[h]
  \caption{Endpoint stability on the pre-frozen 40-task replication subset.}
  \label{tab:supp-stability}
  \centering
  \tablestyle\small
  \setlength{\tabcolsep}{2.5pt}
  \begin{tabular}{@{}>{\raggedright\arraybackslash}p{0.27\linewidth}>{\centering\arraybackslash}p{0.31\linewidth}>{\centering\arraybackslash}p{0.31\linewidth}@{}}
\toprule
Metric & DeepSeek & Grok \\
\midrule
Main subset $\Delta$ & \shortstack[r]{+0.0515{}[-0.0409, 0.1453]} & \shortstack[r]{+0.0517{}[-0.0631, 0.1663]} \\
\addlinespace[2pt]
Replication $\Delta$ & \shortstack[r]{+0.0615{}[-0.0200, 0.1452]} & \shortstack[r]{+0.1414{}[0.0363, 0.2497]} \\
\addlinespace[2pt]
Rep.--main & \shortstack[r]{+0.0100{}[-0.1084, 0.1304]} & \shortstack[r]{+0.0897{}[-0.0333, 0.2316]} \\
\addlinespace[2pt]
Task-$\Delta$ $r$ & 0.076 & 0.257 \\
MAE & 0.2688 & 0.2436 \\
\bottomrule
\end{tabular}
  \smallskip
  \begin{flushleft}
    \scriptsize\textit{Note.} Brackets report 95\% CIs. $\Delta$ is TSM minus ReAct. Rep.--main compares the replication effect with the paired effect on the same 40-task main-run subset.
  \end{flushleft}
\end{table}

%% file: tables/supplementary_provenance_stress.tex
  \captionof{table}{Provenance stress test on 40 injected DeepSeek tasks.}
  \label{tab:supp-provenance-stress}
  \centering
  \tablestyle
  \setlength{\tabcolsep}{3pt}
  \begin{tabular*}{\linewidth}{@{\extracolsep{\fill}}lrrrr@{}}
    \toprule
    Condition & Score & Complete & Seen & Non-active \\
    \midrule
    ReAct & 0.3782 & 38/40 & -- & 0/40 \\
    TSM & \tableemph{0.4074} & 38/40 & 30/40 & \tableemph{38/40} \\
    \bottomrule
  \end{tabular*}
  \smallskip
  \begin{minipage}{\linewidth}
    \footnotesize\raggedright
    \textit{Note.} Four error types were used (10 tasks each): wrong file declaration, cross-city scope, stale version, and missing metadata. Seen counts runs that requested the affected response; Non-active counts runs with a stale or rejected assertion.
  \end{minipage}